\documentclass[10pt,journal,compsoc]{IEEEtran}
\usepackage[accsupp]{axessibility}
\usepackage{times}
\usepackage{epsfig}
\usepackage{graphicx}
\usepackage{svg}
\usepackage{amsmath}
\usepackage{amssymb}
\usepackage{multirow}
\usepackage{array}
\usepackage{tabularx}
\usepackage{caption}
\usepackage{subcaption}
\usepackage{hyperref}
\usepackage{float}
\usepackage{booktabs}
\usepackage{comment}
\usepackage{subcaption}
\usepackage[table]{xcolor}
\hypersetup{
    colorlinks=true,
    linkcolor=blue,
    filecolor=magenta,      
    urlcolor=cyan,
    pdftitle={Overleaf Example},
    pdfpagemode=FullScreen,
    }

\newcolumntype{P}[1]{>{\centering\arraybackslash}p{#1}}

\begin{document}

\title{Do You See What I Say? Generalizable Deepfake Detection based on Visual Speech Recognition}

\author{Maheswar Bora, Tashvik Dhamija, Shukesh Reddy, Baptiste Chopin, Pranav Balaji,\\ Abhijit Das,~\IEEEmembership{Senior Member ~IEEE, Antitza Dantcheva~\IEEEmembership{Senior Member ~IEEE}}
        
        
\IEEEcompsocitemizethanks{
\IEEEcompsocthanksitem M.Bora, S.Reddy, A.Das are with the Machine Intelligence Group, Department of CS\&IS, Birla Institute of Technology and Sciences, Pilani,
India.\protect\\
\IEEEcompsocthanksitem A Dantcheva, B Chopin, T Dhamija, P Balaji, are with STARS team Inria Center at Université Côte d'Azur in Sophia Antipolis, France.\\
E-mail: antitza.dantcheva@inria.fr, abhijit.das@hyderabad.bits-pilani.ac.in.

}

}

%
%

\markboth{JOURNAL OF \LaTeX~CLASS FILES, VOL. 14, NO. 8, AUGUST 2015}%
{M.Bora \MakeLowercase{\textit{et al.}}: Do You See What I Say? Generalizable
Deepfake Detection based on Visual Speech
Recognition}

\IEEEtitleabstractindextext{
\begin{abstract}
Deepfake generation has witnessed remarkable progress, contributing to highly realistic generated images, videos, and audio. While technically intriguing, such progress has raised serious concerns related to the misuse of manipulated media. To mitigate such misuse, robust and reliable deepfake detection is urgently needed.  
Towards this, we propose a novel network FauxNet, which is based on pre-trained Visual Speech Recognition (VSR) features. By extracting temporal VSR features from videos, we identify and segregate real videos from manipulated ones. The holy grail in this context has to do with zero-shot detection, \textit{i.e.,} generalizable detection, which we focus on in this work. 
FauxNet consistently outperforms the state-of-the-art in this setting. 
In addition, FauxNet is able to {attribute} - distinguish between generation techniques from which the videos stem. 
Finally, we propose {new datasets, referred to as Authentica-Vox and Authentica-HDTF, comprising about 38,000 real and fake videos in total}, the latter created with six recent deepfake generation techniques. We provide extensive analysis and results on the Authentica datasets and FaceForensics++, demonstrating the superiority of FauxNet. 
{The Authentica datasets will be made publicly available.}\\
\end{abstract}


}

\maketitle

\vspace{-24mm}
\section{Introduction}
\IEEEPARstart{D}{eepfakes}  consist of images, videos, and audio that have been manipulated by deep neural networks. 
Such manipulations aim to depict individuals, realistically saying or performing actions that have never occurred. 
The rise of deepfakes raises significant ethical concerns, particularly \textit{w.r.t.} misinformation, privacy, and consent. Very recent deepfakes highlight this, focusing on the US elections\footnote{https://www.latimes.com/california/story/2024-08-31/california-is-racing-to-combat-deepfakes-ahead-of-the-election}, as well as on defaming and exploiting the prominence of Taylor Swift\footnote{https://www.bbc.com/news/technology-68110476}.  
Motivated by the above, it is imperative to combat and mitigate the potential misuse of deepfakes by proposing robust and reliable detection approaches. While such detection approaches are advancing, significantly larger efforts have been placed on proposing novel adversary approaches - namely generative models \cite{pei2024deepfakegenerationdetectionbenchmark} fueled by versatile commercial applications. 
Examples include Large Language Models (LLMs) \cite{openai2024gpt4technicalreport, geminiteam2024gemini}, image generation \cite{rombach2022high,ho2022cascaded}, video generation \cite{ho2022imagen,jiang2023text2performer}, as well as face animation in both 3D \cite{Thambiraja_2023_ICCV} and 2D \cite{Drobyshev_2024_CVPR}. 

In this work, we focus on deepfake videos where most recent generative models are able to generate highly realistic facial videos - deepfakes, posing significant challenges for both, computer vision algorithms and humans in distinguishing them from real videos.  
Deepfake generation methods can be categorized into two subsets: (a) video-driven methods \cite{Siarohin2019,wang2022latent,weng2020,av2024latent} that transfer motion patterns from one video depicting an identity to another, as well as 
(b) audio-driven methods \cite{VASA2023,EMO2023,dreamtalker2023} that animate talking heads based on driving audio-speech or textual descriptions. 
Both categories, (a) and (b) 
exhibit the challenge of synchronizing what is being said with lip motion. The latter is challenging to model, as it intricately combines high and low-frequency motion that must be correlated to an audio sequence \textit{w.r.t.} both {types of features}, \textit{temporal}: mouth opening with the right timing, as well as \textit{spatial}: mouth shape corresponding to the sound. 
Therefore, most recent deepfake detection methods have exploited this challenge by analyzing the lip region \cite{9980296,9298826} and have proceeded to extract related features. 

Motivated by the above {challenge}, we adopt a lip-reading method that decodes lip motion to obtain a text transcript of the speech, namely the Visual Speech Recognition (VSR) network.
Specifically, we showcase that video embedding from the VSR Encoder contains pertinent clues for detecting generated videos. Based on this, we design a novel framework for deepfake detection, namely FauxNet that is notably able to (i) distinguish between real and fake videos and moreover to (ii) {attribute and therefore} classify generation approaches.  
We place emphasis on the fact that our method \textit{generalizes to unseen manipulation techniques} and therefore performs on the challenging and pertinent setting of zero-shot deepfake detection. In this setting, FauxNet outperforms the state-of-the-art ({SOTA}).
We note that zero-shot detection represents the holy grail of deepfake detection, given 
the rapid advent of new generative models, which quickly render supervised learning deepfake detection obsolete.\\

\noindent Our main contributions in this paper include the following.\begin{itemize} 
\item We propose a \textit{new dataset, Authentica}, including  two versions based on VoxCeleb2 and HDTF, comprising {about 38,000} videos that were generated by six recent generation techniques. 
\item We introduce a novel \textit{zero-shot multitask framework for deepfake detection} referred to as \textit{FauxNet}, based on VSR features. 

\item We conduct large-scale experiments that suggest our 
FauxNet \textit{outperforms {SOTA}} in the challenging \textit{zero-shot detection setting}. 
In addition, FauxNet is able to \textit{classify} 
\textit{deepfake generation techniques}.
\end{itemize}
\begin{figure}[h!]
    \centering
    \includegraphics[width=1.0\linewidth]{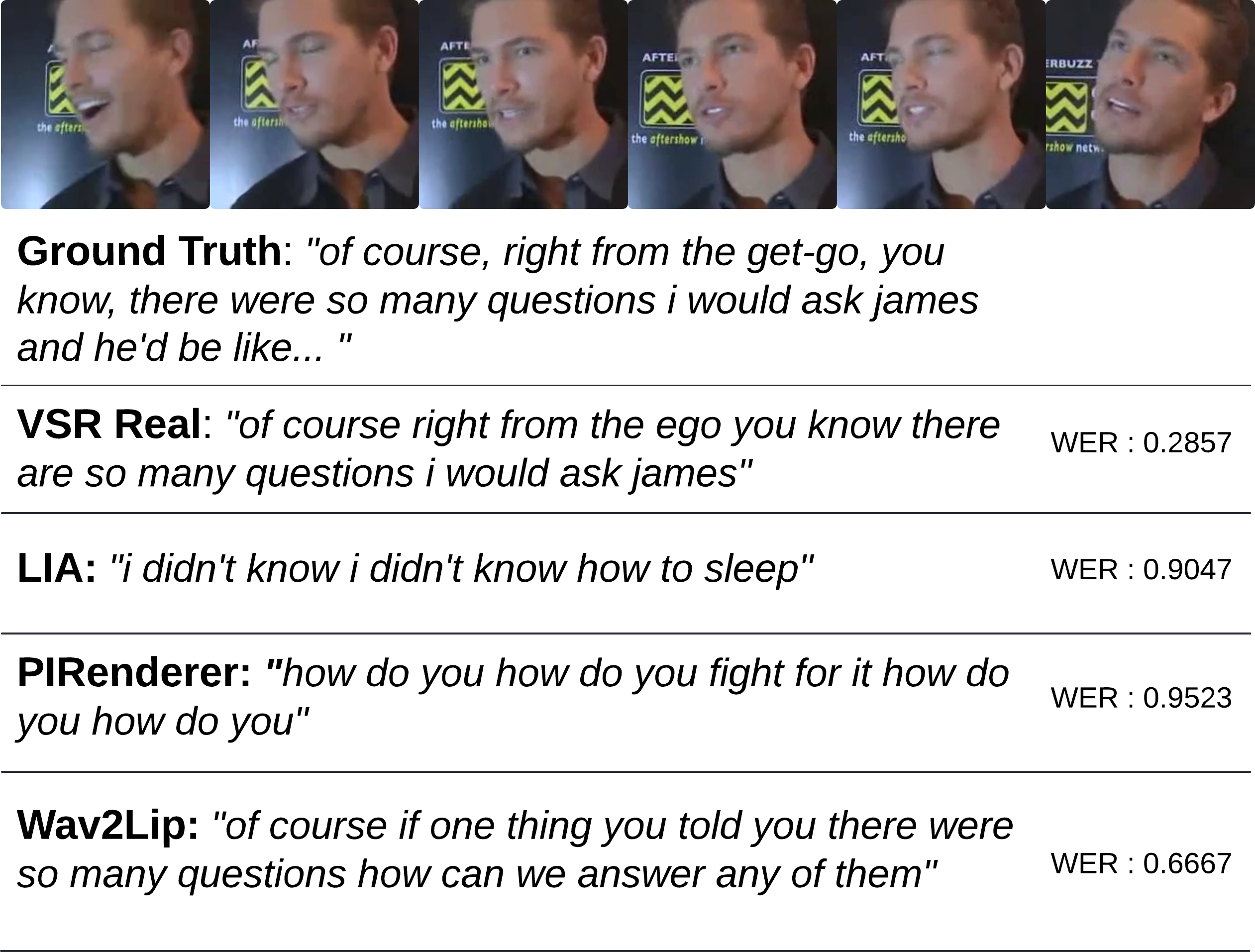}
    \caption{The images shown are the 0th, 20th, 40th, 60th, 80th, and 100th frames (from the left) for the same real video ID. Here, \textbf{Ground Truth} refers to the text extracted from whisper \cite{radford2023robust}  and the remaining labels refer to text extracted using VSR. The Word Error Rate (WER) \cite{morris2004wer} has been calculated against the ground truth text after removing the punctuation marks.}
    \label{fig_ImageTranscript}
\end{figure}
\section{Related Works}
We proceed to revisit the {SOTA} \textit{w.r.t.} contributions in this paper. Specifically, we discuss the following deepfake datasets, which we summarize in Table \ref{tab:Authentica_datastats}, elaborate on the {SOTA} pertained to deepfake detection, classification, as well as VSR, which FauxNet is based on.

\subsection{Deepfake Datasets}
The emergence of deepfakes has motivated the creation of several large-scale datasets to facilitate detection research. The FaceForensics++ dataset \cite{rossler2019faceforensics++} is prominent, consisting of both, real and manipulated videos generated using four different techniques. Similarly, the DeepFake Detection Challenge Dataset (DFDC) \cite{dolhansky2019deepfake} includes millions of frames with various deepfake manipulations, focusing on enhancing model generalization. Celeb-DF \cite{li2020celebdf} also provides vast collections of real and manipulated videos, with an emphasis on high-quality facial reenactment. While these datasets have contributed significantly to detection efforts, they mostly focus on specific manipulation techniques, thus lacking diversity in deepfake generation techniques. Proprietary or novel generative techniques often remain unrepresented, which poses challenges for models to generalize in zero-shot settings. This brings about the need for more comprehensive datasets that include both, video-driven and audio-driven deepfakes, as different generators may exploit different aspects of the data, such as facial texture or lip synchronization.
\renewcommand{\arraystretch}{1.1}
\begin{table*}[!ht]
    \centering
    \caption{{Comparing the proposed dataset, Authentica-Vox, and Authentica-HTDF with existing deepfake detection datasets.}}
    \resizebox{\textwidth}{!}{
    \begin{tabularx}{\textwidth}{|c|c|>{\centering\arraybackslash}X|>{\centering\arraybackslash}X|c|c|c|}
    \hline
        \textbf{Dataset} & \textbf{Year} & \textbf{Audio driven Techniques} & \textbf{Video driven Techniques} & \textbf{Real} & \textbf{Fake} & \textbf{Unique IDs} \\ 
        \hline
        UADFV \cite{li2018exposing} & 2018 & None & FakeApp \cite{fakeapp2019} & 49 & 49 & 49\\
        \hline
        DeepfakeTIMIT \cite{Korshunov2018} & 2018 & None & FaceSwap GAN \cite{shaoanlu_faceswapgan} & 640 & 320 & 32\\
        \hline        FaceForensics++ (FF++)~\cite{rossler2019faceforensics++} & 2019 & None & FaceSwap \cite{FaceswapSoftware},  DeepFakes~\cite{Korshunov2018},  NeuralTextures~\cite{neural_textures_2020},  Face2Face~\cite{face2face_2018} & 1,000 & 4,000 & 1,000 \\
        \hline
        Celeb-DF \cite{li2020celebdf} & 2019 & None & DFVAE-HQ \cite{li2020celebdf} & 590 & 5,639 & 59\\
        \hline
        Google DFD \cite{googledfd2019} & 2019 & None & Undisclosed & 0 & 3,000 & 28\\
        \hline
        DeeperForensics \cite{jiang2020deeperforensics} & 2020 & None & DFVAE \cite{jiang2020deeperforensics} & 50,000 & 10,000 & 100\\
        \hline
        DFDC \cite{dolhansky2019deepfake} & 2019 & None & DFAE \cite{Liu2020DeepfacelabIF}, 
        MM/NN face swap\cite{Huang2012Facial},
        NTH \cite{zakharov2019few}, FSGAN \cite{Nirkin2019FSGAN},
        StyleGAN based \cite{Karras2019Style} & 23,654 & 104,500 & 960 \\
        \hline
        KoDF \cite{Kwon2021KoDFAL} & 2021 & ATFHP \cite{Yi2020AudiodrivenTF}, 
        Wav2Lip \cite{prajwal2020lipsync} & FaceSwap (software) \cite{FaceswapSoftware},  DeepFaceLab \cite{Liu2020DeepfacelabIF},  FSGAN \cite{Nirkin2019FSGAN} & 62,166 & 175,776 & 403 \\
        \hline
        FakeAVCeleb \cite{Khalid2021FakeAVCelebAN} & 2021 & Wav2Lip \cite{prajwal2020lipsync} & Fast FaceSwap \cite{Korshunova2016FastFU},  FSGAN \cite{Nirkin2019FSGAN} &
        500 & 19,500 & 500\\
        \hline
        INDIFACE \cite{kuckreja2024indiface} & 2024 & None &
        SimSwap \cite{10.1145/3394171.3413630},  Ghost \cite{9851423} & 404 & 1668 & 58 \\
        \hline
        \hline
        \textbf{Authentica-Vox (Ours)} & \textbf{2024} & \textbf{StyleTalk~\cite{StyleTalk2022},  DreamTalk~\cite{dreamtalker2023}, 
        SadTalker ~\cite{Chen2023},
        Wav2Lip~\cite{prajwal2020lipsync}} & \textbf{LIA~\cite{wang2022latent},PiRenderer~\cite{conf/iccv/Ren0CL021}} & \textbf{3,916} & \textbf{31,326} & \textbf{37} \\ 
        \hline
        \textbf{Authentica-HDTF (Ours)} & \textbf{2024} & \textbf{StyleTalk~\cite{StyleTalk2022},  DreamTalk~\cite{dreamtalker2023}, 
        SadTalker ~\cite{Chen2023},
        Wav2Lip~\cite{prajwal2020lipsync}} & \textbf{LIA~\cite{wang2022latent},PiRenderer~\cite{conf/iccv/Ren0CL021}} & \textbf{409} & \textbf{2,744} & \textbf{265} \\ 
        \hline
    \end{tabularx}
    }
    
    \label{tab:Authentica_datastats}
\end{table*}
\subsection{Deepfake Detection}
Numerous deepfake detection methods have been proposed to tackle the rising prevalence of synthetic videos. The Two-Stream Network \cite{zhou2017two,das-limiteddatavit-wacv2024,reddyICPR24} leverages both, spatial and temporal information, to detect forgeries. However, it is limited as it overfits to known manipulation techniques. Capsule Networks \cite{nguyen2019capsule} have been used to address forgery detection by capturing spatial relationships between facial landmarks. XceptionNet \cite{rossler2019faceforensics++}, which leverages the Xception architecture, has become a popular choice for detecting pixel-level anomalies in manipulated videos. EfficientNetB4 \cite{9412711} proposes an ensemble of EfficientNets to perform deepfake detection. However, these methods are evaluated on test sets having a similar distribution as the training data. There is a lack of experimentation on a zero-shot framework, especially on recently proposed generative techniques. Even though LAA-Net \cite{Nguyen_2024_CVPR} proposes a model capable of cross-dataset testing, the datasets used do not fairly represent the recently proposed techniques.
The key intuition is that these detection methods are not designed to handle advanced generative techniques and lack generalization over newly proposed generative techniques. As a result, these methods struggle to perform effectively in out-of-distribution scenarios. This capability is increasingly critical, given the rapid evolution of generative models, which produce a wide range of manipulations.
\subsubsection{Zero-Shot Detection}
Zero-shot deepfake detection remains a significant challenge, particularly as generative models rapidly evolve and introduce new manipulation techniques. Traditional detection methods often rely on extensive training data that encompasses various generative techniques; however, these approaches can falter when confronted with unseen manipulations. For instance, work by Yang et al. \cite{yang2020detection} demonstrated that ensemble learning techniques could enhance detection performance, yet they still required exposure to a range of forgery techniques during training. More recent efforts, such as those by Zhang et al. \cite{zhang2020generalizing}, have highlighted the potential of synthesized samples for generalizing to unseen categories, indicating a path forward for zero-shot approaches. Nevertheless, many existing systems overlook the importance of lip artifacts, which is crucial for accurately identifying manipulated videos, particularly those generated through audio-driven techniques \cite{bohacek2024lost}. Therefore, there is an imminent need for frameworks that operate in zero-shot detection.
\subsubsection{Lip-Based Deepfake Detection}
Deepfake detection methods that focus on the lip region have gained traction due to the inherent challenges in accurately modeling lip synchronization in generated videos. For instance, LipForensics \cite{9962982} extracts features related to lip motion to classify fake videos, while other methods, such as the one proposed by Shahzad et al. \cite{9980296}, leverage multimodal features for forgery detection. Additionally, Yang et al. \cite{9298826} analyze dynamic lip movements to enhance the robustness of speaker authentication systems against deepfake attacks. However, these models often learn lip features that become signatures of the specific training datasets, limiting their ability to generalize across different deepfake techniques. This is particularly concerning as manipulation techniques evolve and diversify. 
\subsection{Deepfake Classification}
Classifying the specific techniques used to generate deepfakes is an under-explored area. 
As generation techniques become increasingly diverse, it is essential to not only detect manipulations but also accurately {attribute -} classify the generative technique employed. Despite some advancements, such as those proposed by Arshed et al. \cite{arshed2024multiclass}, there remains a significant gap in the literature regarding comprehensive approaches to deepfake attribution, underscoring the need for more robust methodologies that can generalize effectively to novel and proprietary techniques.
\subsection{Visual Speech Recognition (VSR)}
VSR decodes lip movements into text and has emerged as a viable tool for {lip reading, as well as} deepfake detection, particularly where lip motion is critical. Various models have been developed for VSR, such as LipNet \cite{assael2017lipnet}, which uses convolutional networks to predict phonemes, and methods leveraging 3D CNNs \cite{chen2016lipreading} and Transformer architectures \cite{zhu2020visual} that enhance temporal feature extraction. Recent advancements, including Auto-AVSR \cite{ma2023auto}, have focused on automatic labeling to improve model efficiency. Research indicates that VSR systems struggle with manipulated videos, as highlighted by Bohacek and Farid \cite{bohacek2024lost}, revealing their potential for detecting deepfakes. 

\begin{figure*}[h!]
    \centering
    \includegraphics[width=0.8\linewidth]{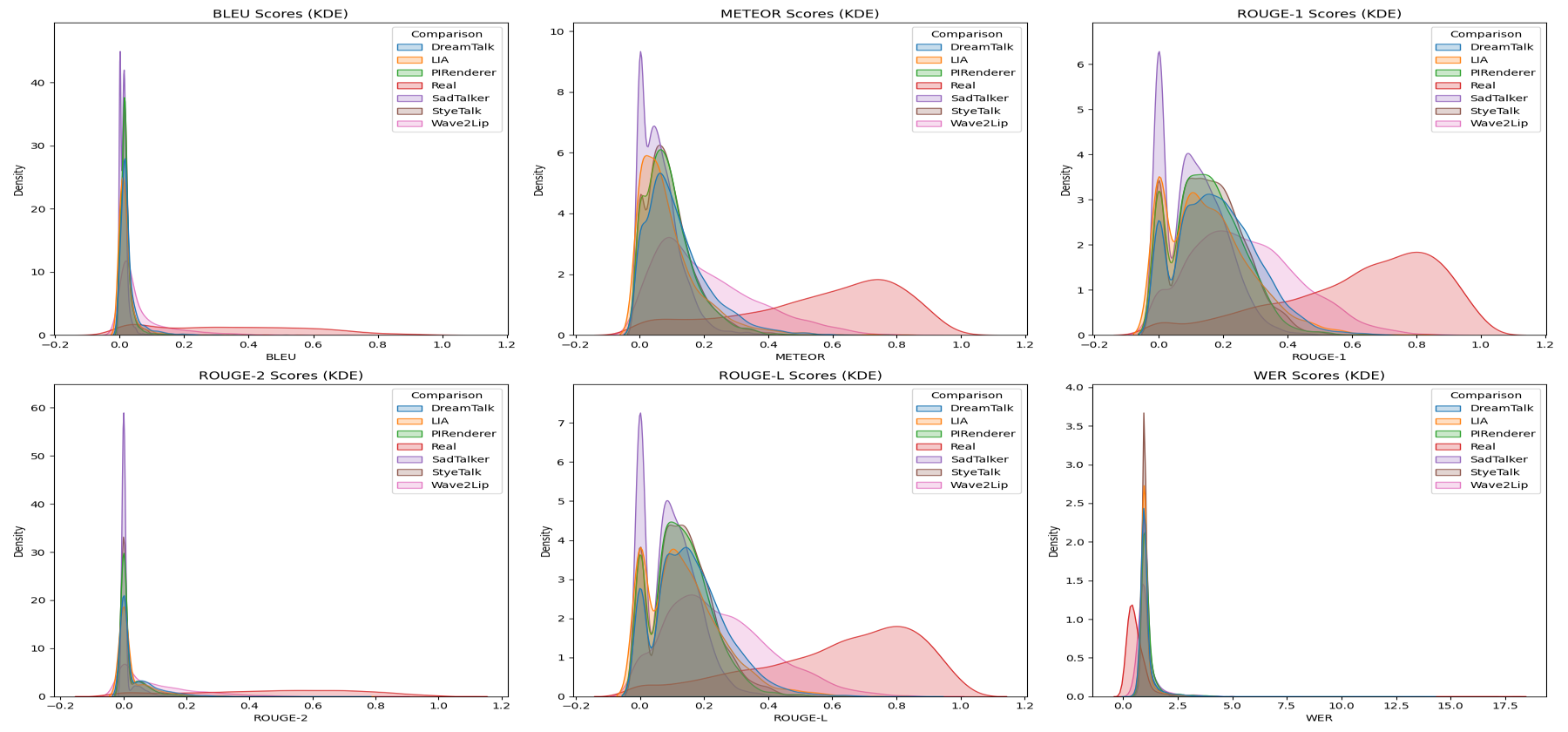}
    \caption{Kernel Density Estimation (KDE) plot for the deviation between Ground truth and VSR transcripts for LIA \cite{wang2022latent}, PiRender-based techniques (PiRenderer \cite{conf/iccv/Ren0CL021}, StyleTalk \cite{StyleTalk2022}, DreamTalk \cite{dreamtalker2023}), SadTalker \cite{Chen2023}, Wav2Lip \cite{prajwal2020lipsync} and real videos. Deviation measured using BLEU \cite{Papineni02bleu:a}, METEOR \cite{lavie2007meteor}, ROUGE-1, ROUGE-2, ROUGE-L \cite{lin-2004-rouge}, WER \cite{morris2004wer} score on proposed Authentica-Vox dataset.}
    \label{fig_vsr_kde_plot_vox}
\end{figure*}


Therefore, while deepfake detection has seen considerable progress, current methods do not generalize to unseen deepfake generation techniques, particularly in zero-shot settings. Moreover, existing approaches focusing on lip-based detection face challenges when both, audio and video are manipulated. These issues are compounded by a lack of a dataset with recent manipulation techniques that represent the real-world problem fairly. Addressing these limitations, we propose a framework that leverages VSR latent encodings to extract visual speech features, allowing for both, robust deepfake detection and classification of the generation technique. We present experiments on Authentica, a new dataset, which we have created {and will release}. We proceed to introduce it.
\section{Authentica: Proposed Dataset{s}}
{In this work, we propose Authentica, a combination of two deepfake datasets, namely Authentica-Vox and Authentica-HDTF. Authentica-Vox consists of 35,242 videos in total including 3,916 real videos from VoxCeleb2 \cite{chung2018voxceleb2} that are used to generate 31,326 fake videos in the proposed dataset. It includes 37 unique identities with videos per identity ranging between 1 to 500 and video duration ranging from 3-20 seconds per video. On the other hand, Authentica-HDTF consists of 2744 videos, of which 409 are real videos from the HDTF~\cite{zhang2021flow} dataset used to generate the fake videos for the proposed dataset. Authentica-HDTF includes 265 unique identities with video duration ranging from 1-4 minutes for each video. By capturing the characteristics of modern deepfake videos, Authentica serves as a valuable resource for advancing research in this area.} Specifically, the datasets include videos pertained to Wav2Lip \cite{prajwal2020lipsync}, LIA \cite{wang2022latent}, PiRenderer \cite{conf/iccv/Ren0CL021}, StyleTalk \cite{StyleTalk2022}, SadTalker \cite{Chen2023}, and DreamTalk \cite{dreamtalker2023} as deepfake generators. 

\noindent\textbf{Latent Image Animator (LIA)} \cite{wang2022latent} is a state-of-the-art model, designed to animate a single image based on the motion of a driving video. It does this by navigating linearly through the motion-latent space, without requiring structural representation. 

\noindent\textbf{PiRenderer} \cite{conf/iccv/Ren0CL021} is a semantic neural rendering model that enables controllable portrait image generation by modifying attributes such as facial expression and head pose, while preserving identity and background elements. It employs a set of motion descriptors derived from a 3D Morphable Model (3DMM) \cite{deng2019accurate} to provide intuitive control over image attributes. 

\noindent\textbf{StyleTalk } \cite{StyleTalk2022} introduces a framework for generating talking head videos that can mimic diverse speaking styles. It utilizes a style encoder to extract dynamic facial motion patterns from a reference video and generates stylized facial animations using a style-controllable decoder. This technique enables the synthesis of talking head videos from a single image and audio clip, capturing the essence of the reference speaking style while maintaining visual authenticity.

\noindent\textbf{SadTalker} \cite{Chen2023} is a talking face generation technique that learns to create realistic 3D motion coefficients for stylized audio-driven talking face animation from a single image. It incorporates full-image animation capabilities and uses pre-trained models to improve the expressiveness and realism of the generated talking face.

\noindent\textbf{DreamTalk} \cite{dreamtalker2023} is a diffusion-based, audio-driven expressive talking head generation framework capable of producing high-quality talking head videos with a variety of speaking styles. It handles diverse inputs, including different languages, songs, noisy audio, and various portrait styles. DreamTalk utilizes diffusion probabilistic models to achieve realistic and emotionally expressive talking face generation from a single image.

\noindent\textbf{Wav2Lip} (or W2L) \cite{prajwal2020lipsync} is a model for lip synchronization in videos, focusing on matching lip movements to audio segments for various identities in unconstrained environments. It learns from a lip-sync discriminator to improve the accuracy of lip movements in talking head videos. The generated lips are then overlayed onto the existing video without tampering with the rest of the video.

For StyleTalk and DreamTalk, each deepfake video is available in two versions: one incorporating head pose motion derived from real video and another one without head pose motion, reflecting the conditioning methods of audio-driven models. 


We compare our proposed Authentica dataset with popular deepfake datasets such as UADFV \cite{li2018exposing}, DeepfakeTIMIT \cite{Korshunov2018}, FaceForensics++~\cite{rossler2019faceforensics++}, Celeb-DF \cite{li2020celebdf}, Google DFD \cite{googledfd2019}, DeeperForensics \cite{jiang2020deeperforensics}, DFDC \cite{dolhansky2019deepfake}, KoDF \cite{Kwon2021KoDFAL}, FakeAVCeleb \cite{Khalid2021FakeAVCelebAN}, and INDIFACE \cite{kuckreja2024indiface} datasets. As {reported in} Table~\ref{tab:Authentica_datastats}, apart from \cite{Kwon2021KoDFAL} and \cite{Khalid2021FakeAVCelebAN}, no other dataset contains audio-driven techniques. Whilst KoDF and FakeAVCeleb contain audio-driven methods, they are limited to primitive early work done in audio-based deepfake generation research available at the time. Unlike the mentioned datasets, we provide a variety of methods using frameworks like latent diffusion \cite{ho2020denoising} in DreamTalk, generative adversarial networks \cite{goodfellow2014generative} in Wav2Lip, and variational autoencoder \cite{kingma2014semi} in Sadtalker. {D}ifferent framework-based methods account for different types of deepfakes {that} a deepfake detection model {is able to} learn to identify. Moreover, most other datasets {are limited} to {the setting of} face-swap whereas, with PiRenderer, we also include 3D motion-driven techniques in our dataset.
\begin{figure*}[htb!]
    \centering
    \begin{subfigure}[b]{0.8\linewidth}
        \centering
        \includegraphics[width=0.8\linewidth]{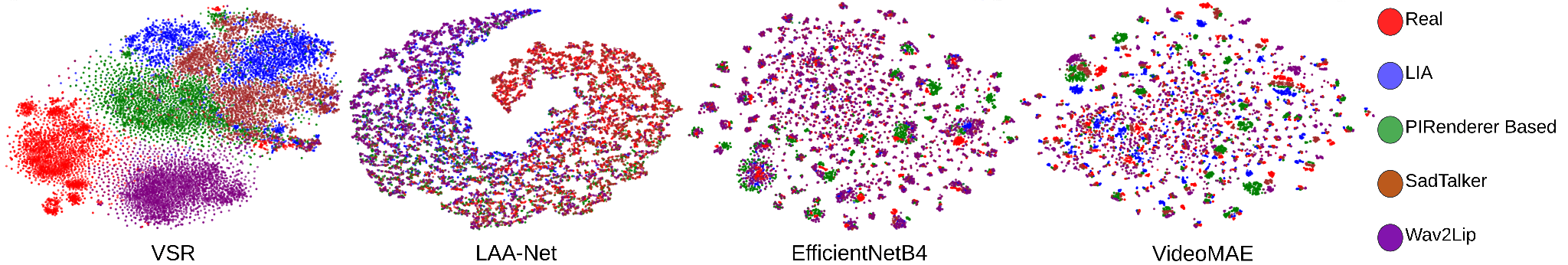}
        \caption{Authentica-Vox}
        \label{fig:tsne_vox}
    \end{subfigure}
    \begin{subfigure}[b]{1.0\linewidth}
        \centering
        \includegraphics[width=1.0\linewidth]{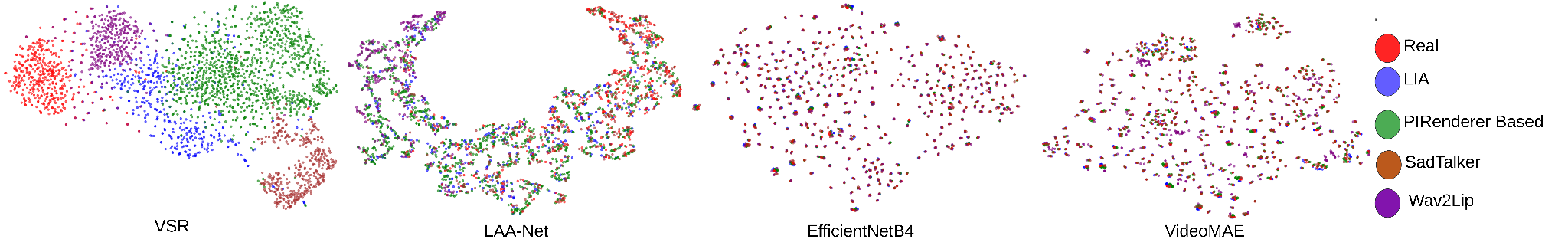}
        \caption{Authentica-HDTF}
        \label{fig:tsne_hdtf}
    \end{subfigure}
    \caption{{T-SNE plots of feature embeddings from the VSR Encoder in different models on the test set of the proposed datasets. (a) corresponds to the proposed Authentica-Vox dataset, while (b) corresponds to the proposed Authentica-HDTF dataset.}}
    \label{fig:combined_tsne}
\end{figure*}
\section{Adopting VSR}
VSR aims to 
interpret lip movements. 
This requires learning the temporal dynamics of lip motion from visual inputs, enabling the model to create a rich feature space that encodes speech information directly from the RGB frames of a video. The temporal aspect of lip movements is essential for capturing the intricacies of speech, and the VSR encoder excels at extracting these temporal features \cite{assael2017lipnet}.

In our methodology, we utilize the Auto-AVSR model~\cite{ma2023auto}, as we identified it to be the most accurate currently available model. Upon testing Auto-AVSR on various samples of real videos from the VoxCeleb dataset~\cite{chung2018voxceleb2}, we observe that the VSR model {is accurate and robust in transcribing} video{s}. However, when the same model is used to visually transcribe a video reconstructed by LIA \cite{wang2022latent}, PiRenderer \cite{conf/iccv/Ren0CL021} and other {generation} techniques, even {with} the same dialogue and visually aligning with the expected lip movement, the VSR model produces transcripts with high error and fails to "understand" the reconstructed deepfakes, see Figure~\ref{fig_ImageTranscript}. 

\begin{figure*}[h!]
    \centering
    \includegraphics[width=0.8\linewidth]{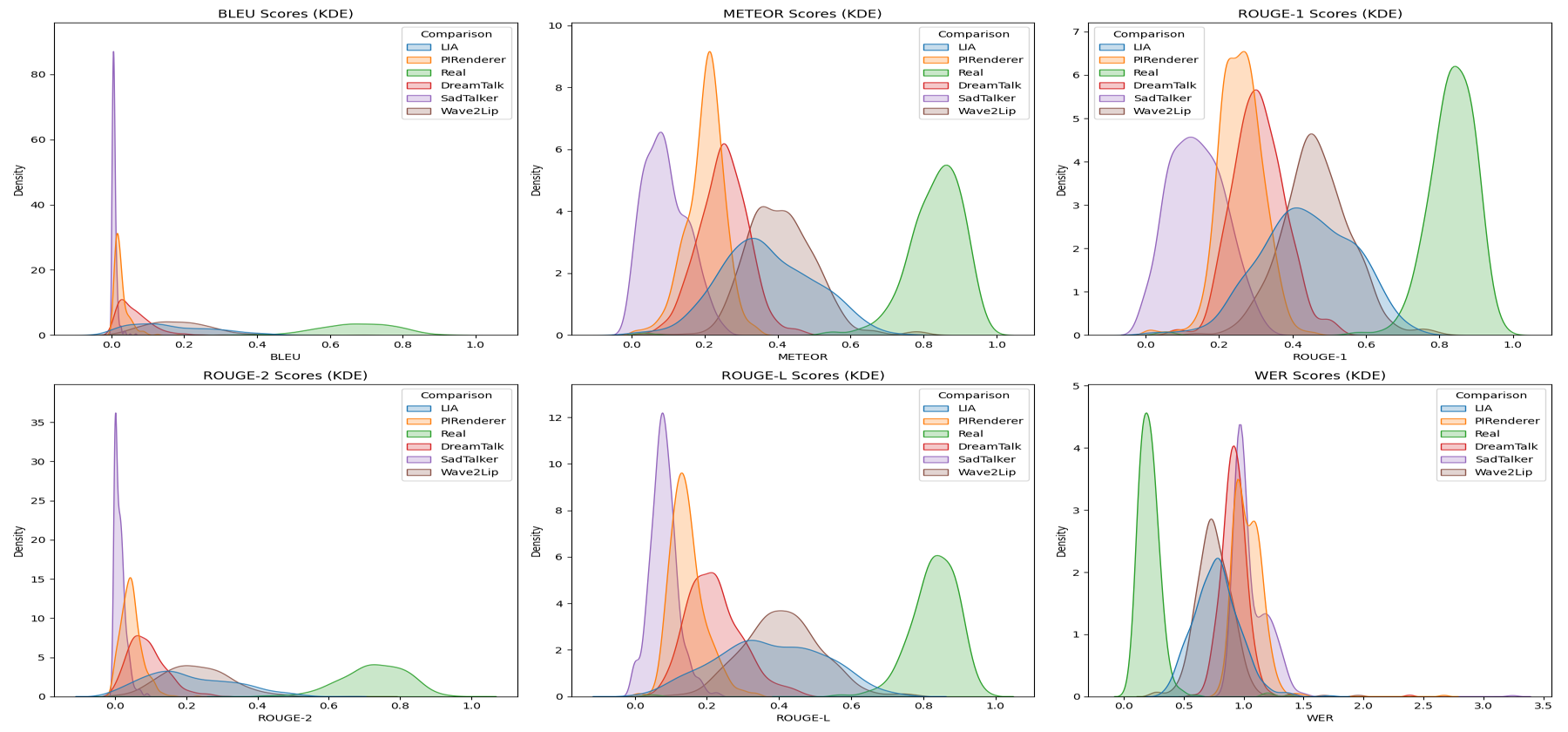}
    \caption{Kernel Density Estimation (KDE) plot for the deviation between Ground truth and VSR transcripts for LIA \cite{wang2022latent}, PiRender-based techniques (PiRenderer \cite{conf/iccv/Ren0CL021}, StyleTalk \cite{StyleTalk2022}, DreamTalk \cite{dreamtalker2023}), SadTalker \cite{Chen2023}, Wav2Lip \cite{prajwal2020lipsync} and real videos. Deviation measured using BLEU \cite{Papineni02bleu:a}, METEOR \cite{lavie2007meteor}, ROUGE-1, ROUGE-2, ROUGE-L \cite{lin-2004-rouge}, WER \cite{morris2004wer} score on proposed Authentica-{HDTF} dataset.}
    \label{fig_vsr_kde_plot_hdtf}
\end{figure*}

Building on this observation, we transcribe real videos, as well as generated videos generated using Wav2Lip \cite{prajwal2020lipsync}, LIA \cite{wang2022latent}, PiRenderer (+ audio-driven techniques using Pirenderer) \cite{conf/iccv/Ren0CL021} and SadTalker \cite{Chen2023} included in the proposed Authentica{-Vox} dataset using the VSR model and calculate the BLEU \cite{Papineni02bleu:a}, METEOR \cite{lavie2007meteor}, ROUGE-1, ROUGE-2, ROUGE-L \cite{lin-2004-rouge}, WER \cite{morris2004wer} score with respect to the ground truth transcript. Given the remarkably high accuracy of audio-based transcription, these transcriptions, extracted from Whisper \cite{radford2023robust} are treated as ground truth. Then, we visualize the kernel density estimation plot obtained for the different metrics in Figure \ref{fig_vsr_kde_plot_vox}. As seen in the Figure, scores significantly deviate between real and the different fake videos. However, when analyzed qualitatively, the lip movement of the reconstructed videos seemingly matches the speech in the audio. This implies that the VSR encoder produces noisy features when attempting to transcribe a deepfake, even though it may not be visually evident. {The same pattern can be observed for the Authentica-HDTF test set as well (see Figure \ref{fig_vsr_kde_plot_hdtf}).} 
Based on this observation, we employ the VSR Encoder model as our feature extractor.

\begin{figure*}[t!]
    \centering
    \includegraphics[width=.80\linewidth]{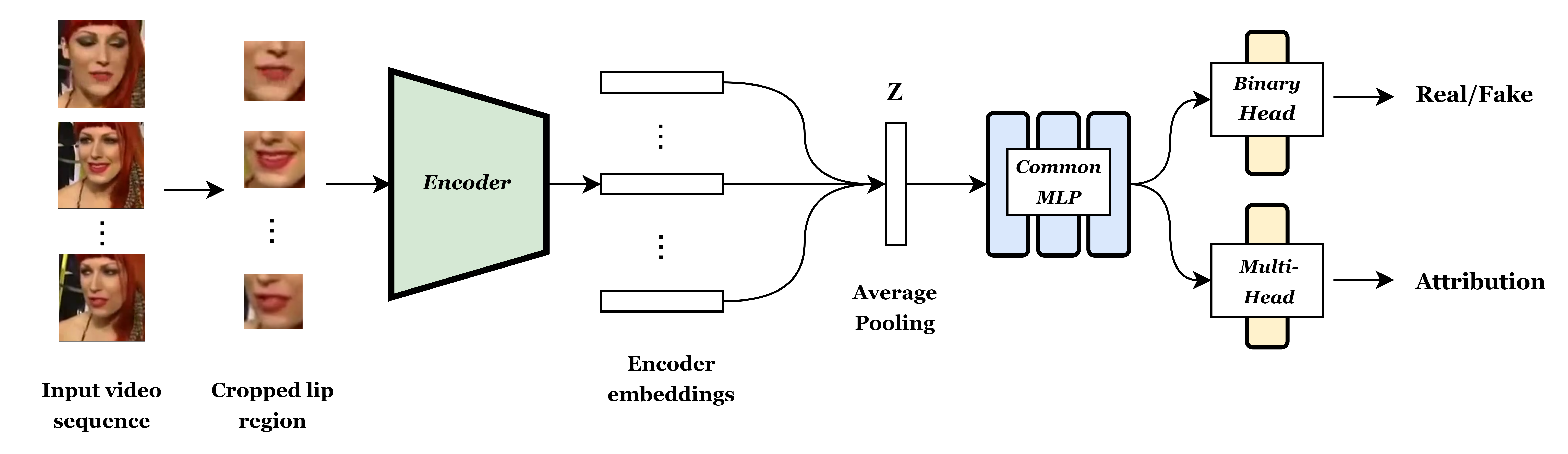}
    \caption{Deepfake detection and classification of deepfake generation technique. We crop the lip region and provide it to VSR as input. Then, the embeddings generated by the VSR-encoder are averaged pooled along the time dimension to obtain one unified video embedding $Z$. This is passed into the common MLP to obtain $Z_c$, which is then used by the two linear heads for detection (\textit{i.e.,} real/fake) and classification (\textit{i.e.,} which type of manipulation). }
    \label{vsr_main_pipeline}
\end{figure*}

\section{Proposed FauxNet}
Based on the analysis of the previous section, we propose a deepfake detection model referred to as FauxNet. In our pipeline, we first crop the lip region in the video based on the facial landmarks \cite{bulat2017far}. This sequence of lip frames is next passed through the VSR encoder, in order to extract features in a windowed manner. For a video \( X \) of length \( T \), the window is defined as \( X \in \mathbb{R}^{H \times W \times 3 \times w} \), where \( w\) is the window length of the VSR encoder and \( d\) represents the dimension size of the VSR encoder for each timestep. This is encoded as 
$Z^{1:w} = \text{VSR\_encoder}(X^{1:w})$, where
$Z^{1:w} \in \mathbb{R}^{w \times d}$.
In this context, \(w\) denotes the window size, and \(d\) is the VSR Encoder's embedding dimension. Once the temporal feature sequence is obtained for the whole video, it is aggregated across the time domain to obtain a single time step representation for a video. In our work, we use average pooling as our temporal sequence downsampler. This process is run over the entire train set,
\[
Z = \text{AveragePool}(Z^{1:w}) = \frac{1}{w} \sum_{i=1}^{w} Z^{i}.
\]
The features extracted from the training set are used to train a classification head that distinguishes real videos from fake ones. 
\section{{Attribution of Deepfakes}}
Figure \ref{fig_vsr_kde_plot_hdtf} illustrates that feature representations of real videos substantially deviate from those of fake videos. To build further intuition, we visualize the VSR encoder's latent space. To do so, we extract the features from real and fake videos and plot the t-SNE \cite{vandermaaten2008tsne} distribution of all samples in the test set {of both subsets} in the proposed Authentica dataset (see Figure~\ref{fig:combined_tsne}). In particular, samples from different generative models cluster distinctively in the latent space. This indicates the ability of the VSR encoder to provide features that are instrumental in {attributing deepfakes}. Consequently, we use a set of multilayer perceptrons (MLP) to cluster different types of generative techniques, in order to classify deepfake generation techniques.

\subsection{Multi Task Learning}
We aim to perform two separate, yet mutually dependent, tasks of deepfake detection and {attribution}. Therefore, we formulate our classification head as a multi-task learning problem. Rather than treating a problem as an auxiliary task, we consider both tasks equally important and train them jointly. We visualize the model in Figure~\ref{vsr_main_pipeline}. Once the extracted features are average pooled, a common MLP projects the pooled features. These features are passed to the BinaryHead for detection and the MultiHead for classification of deepfake generation techniques, only in the case of fake samples. For input \(Z\), the ground truth detection label \(y_{dd}\) takes on the value 0 for real and 1 for fake, whereas \(y_{dt}\) is the deepfake generative technique label for a fake sample. The loss used to train the model \(L_{\text{total}}\) is defined as follows.
\[Z_c = \text{CommonMLP}(Z),\]
\[\hat{y}_{dd} = \text{BinaryHead}(Z_c),\]
\[\hat{y}_{dt} = \text{MultiHead}(Z_c) \quad \text{if } y_{dd} = 1,\]
\[L_{\text{bce}} = \text{BinaryCrossEntropy}(\hat{y}_{dd}, y_{dd}),\]
\[L_{\text{ce}} = \text{CrossEntropy}(\hat{y}_{dt}, y_{dt}),\]
\[L_{\text{total}} = L_{\text{bce}} + y_{dd}  L_{\text{ce}}.\]
\section{VSR text-based clustering}
Building on the intuition provided by {Bohacek and Farid} \cite{bohacek2024lost}, we compare our proposed model with an end-to-end VSR text clustering model. 
For classification using decoded VSR text, we evaluate real and fake transcripts against the ground truth text that was extracted by Whisper \cite{radford2023robust}. We calculate six metrics - BLEU \cite{Papineni02bleu:a}, METEOR\cite{lavie2007meteor}, ROUGE-1, ROUGE-2, ROUGE-L \cite{lin-2004-rouge}, as well as WER \cite{morris2004wer} score. These metrics are normalized to range between 0 and 1. We find the best threshold, separating real and fake transcript text of the training set. The threshold is calculated individually for each metric. The final accuracy on the test set is computed by majority voting of the predictions of the six metrics. 


\section{Experiments}
\subsection{Experimental Setup}
In this work, we evaluate our proposed method on FaceForensics++ \cite{rossler2019faceforensics++}, as well as our proposed Authentica dataset. For deepfake detection, we compare our model with the {SOTA}, LAA-Net \cite{Nguyen_2024_CVPR} and EfficientNetB4 \cite{9412711}. Additionally, we compare our model with a VideoMAE \cite{tong2022videomae} backbone classification model along with LAA-Net and EfficientNetB4. 

\noindent\textbf{General setup.} In our experimentation, we follow preprocessing protocols, as originally proposed for the respective baseline models. As mentioned in the methodology, we obtain latent features from the pre-trained VSR encoder \cite{ma2023auto} trained on 3,448 hours of video data. The lip region of the dataset is extracted via cropping and resizing to a dimension of 96x96 pixels following the protocol mentioned in \cite{ma2023auto}. The embeddings after the VSR encoder are averaged along the sequence dimension, in order to obtain a single video embedding $Z_c$ of size 768. The MLP classifies this embedding. 
For the EfficientNet \cite{9412711} feature extractor, we use the model released by the authors which were pre-trained on FaceForensics++ \cite{rossler2019faceforensics++}. The 32 evenly spaced frames are extracted from the video and resized to a size of 224x224 pixels. We then average the frame embeddings towards obtaining a single video representation. For LAA-Net \cite{Nguyen_2024_CVPR}, we use the pre-trained FaceForensics++ \cite{rossler2019faceforensics++}  model. 32 evenly spaced frames from the video are taken, cropped, and resized to 384x384 pixels. The frame embeddings are averaged, resulting in a single video representation. Similarly for VideoMAE \cite{tong2022videomae}, we use the pre-trained large variant of the model, trained on Kinetics-400 \cite{kay2017kinetics}, for feature extraction on frames of 224x224 dimension.
As a precaution check, for samples with fewer than 32 frames, we duplicate the last frame till we reach 32 frames. However, we have ensured to exclude such samples from the Authentica dataset.
\\
\newline{}
\textbf{Authentica-HDTF setup.} For the experiments using Authentica-HDTF, we create 15-second non-overlapping chunks of the videos due to the large duration (1-4 minutes) of the videos. For Authentica-Vox and FaceForensics++ \cite{rossler2019faceforensics++} we utilize frames only up to the first 15 seconds since the videos are usually not longer than 10-20 seconds. When used for training and validation, we use all the extracted chunks, since the encoders are unable to perform well beyond 15 seconds. When used for testing we utilize the first chunk i.e., the first 15 seconds only. For fairness, we utilize the first 32 frames of the video for the FauxNet experiments. The VSR encoder of FauxNet requires the frames to be consecutive; hence, we take the first 32 frames rather than evenly spaced 32 frames. Optionally, any consecutive 32-frame sequence could have been taken. For the One-vs-All testing experiment, we use only Authentica-HDTF and have created a 70\%-10\%-20\% train-validation-test split. We have ensured that for all experiments the identities are not shared between the splits. Furthermore, the identities are disjoint across the datasets. For the Authentica-Vox experiments, we create a 80\%-10\%-10\% train-validation-test split. The larger test set ratio for Authentica-HDTF was to increase the diversity and number of samples of the test set. 
\\
\newline{}
\textbf{CommonMLP.} The CommonMLP consists of 3 hidden layers with the 1D batch norm, ReLU \cite{nair2010rectified} activation, and dropout with a probability of 0.5 after every layer. For BinaryHead and MultiHead, we use a single linear layer. For the binary classification task \textit{i.e.,} the real versus fake classification task, we use sigmoid activation followed by binary cross-entropy loss. For the multi-class deepfake generator detection task we use softmax activation followed by cross-entropy loss. Finally, we test the models on the checkpoint, giving the lowest validation loss.
\\
\newline{}
 \textbf{Training details.} To train the model we use the AdamW \cite{Loshchilov2017DecoupledWD} optimizer with a learning rate of 0.0005 and weight decay regularization value of 0.00001. The learning rate is reduced by a factor of 0.5 if the validation loss does not improve for 3 consecutive epochs. Furthermore, early stopping is applied, given that the validation loss did not improve for 10 consecutive epochs. The MLP is trained with a batch size of 256 for a maximum of 100 epochs on NVIDIA 2080Ti GPUs.

\subsection{Results on in-distribution test set}
We evaluate the in-distribution performance of the proposed FauxNet, LAA-Net \cite{Nguyen_2024_CVPR} and EfficientNetB4 \cite{9412711}. Towards this, we train our model on the FF++ dataset, abiding by the experimental setup of the models to be compared with. From Table \ref{tab:indist}, we observe that our model substantially outperforms EfficientNet \cite{9412711}. We also obtain on-par results with the current {SOTA}, LAA-Net \cite{Nguyen_2024_CVPR}. This showcases the strength of the information available in the pre-trained VSR encoder's features for deepfake detection. Despite the feature extractor never being trained on the FF++ dataset, it can extract features useful for the MLP and Binary head to detect fake videos. This also showcases the robustness of the feature extractor to a shift in the distribution of input and acts as validation for the model to be robust against different kinds of real-world scenarios. {Moreover, to validate our choice of using VSR as our feature extractor, we replace FauxNet's feature extractor with VideoMAE \cite{tong2022videomae}. 
Results in Table \ref{tab:indist} show that despite VideoMAE's benefits, 
the VSR encoder outperforms it, confirming our hypothesis.
}
\begin{table}[h]
    \centering
    \caption{In-distribution deepfake detection. We compare FauxNet and {SOTA} on the FF++ testing set. {Best score is in bold, second best is underlined}.}
    \begin{tabular}{|c|c|c|}
        \hline
        \textbf{Method} & \textbf{AUC} & \textbf{Accuracy\%}  \\
        \hline
        EfficientNet B4 \cite{9412711} & 0.9382 & -\\
        \hline
        LAA-Net \cite{Nguyen_2024_CVPR} & \textbf{0.9996} & - \\
        \hline
        {VideoMAE \cite{tong2022videomae}} & {0.983} & {94.57} \\
        \hline
        \textbf{FauxNet (ours)} & \underline{0.9938} & {\textbf{96.00}} \\
        \hline
        \textbf{FauxNet (ours) (Zero-shot)} & 0.9143 & 84.43 \\
        \hline
    \end{tabular}
    \label{tab:indist}
\end{table}
\subsection{Results on zero-shot evaluation}
To evaluate the zero-shot performance of FauxNet, we conduct two separate experiments. Firstly, we train our model on the proposed Authentica{-Vox} train set and test it on the FF++ dataset. As reported in Table \ref{tab:indist}, our model trained on Authentica provides competitive results to EfficientNetB4 \cite{9412711}, despite neither the feature extractor nor the MLP + Binary head being trained on the FF++ dataset.
Secondly, to fairly compare the zero-shot performance of our model with the rest of the networks, we evaluate all models trained on FF++ on the proposed Authentica{-HDTF} testing set. In addition to the previous SOTA methods, we add here LipForensics \cite{haliassos2021lips} also trained on FF++. Table \ref{tab:ffpp_train_hdtf_ood_test}, summarizes that our model generalizes substantially better to zero-shot samples - without any training or adaptation. This further showcases that {SOTA} is challenged in the setting of out-of-distribution / zero-shot. This lack of generalizability {indicates} that these models are not adequate in real-world scenarios. {In addition, replacing the VSR encoder in FauxNet with the VideoMAE encoder leads to weakened generalization capabilities despite VideoMAE being trained for effective video representation.}
Furthermore, we train our model without multitask learning, \textit{i.e.,} removing MultiHead from the proposed model. Whilst it performs better in the zero-shot setting as compared to other models, we observe how the proposed FauxNet benefits from multitask learning, specifically in case that {attribution} is learned in parallel with deepfake detection.

\begin{table}[h]
    \centering
    \caption{{Zero-shot deepfake detection. We compare FauxNet and SOTA trained on FF++ in zero-shot testing on Authentica-HDTF.}}
    \begin{tabular}{|c|c|c|}
        \hline
        \textbf{Method} & \textbf{AUC} & \textbf{Accuracy\%} \\
        \hline
        Efficient B4 \cite{9412711} & 0.7555 & 32.47 \\
        \hline
        LAA-Net \cite{Nguyen_2024_CVPR}  & 0.508 & 16.44 \\
        \hline
        LipForensics \cite{haliassos2021lips} & 0.5963 & 48.00\\
        \hline
        VideoMAE \cite{tong2022videomae} (w/o multi-task) & 0.7343 & 84.11 \\
        \hline
        VideoMAE \cite{tong2022videomae} (multi-task) & 0.7127 & 85.20 \\
        \hline
        FauxNet (ours) (w/o multi-task)  & 0.9962 & 88.93 \\
        \hline
        \textbf{FauxNet (ours)}  & \textbf{0.9969} & \textbf{92.58} \\ 
        \hline
    \end{tabular}
    \label{tab:ffpp_train_hdtf_ood_test}
\end{table}

\subsection{Results on {attribution of deepfakes}}
\begin{figure*}[h!]
    \centering
    \includegraphics[width=1.0\linewidth]{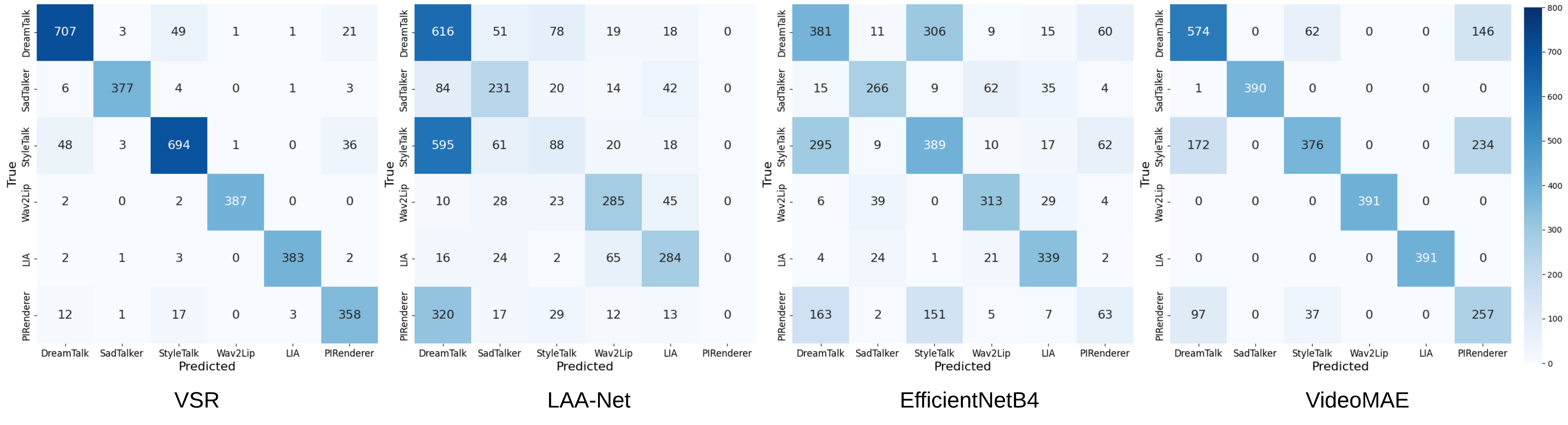}
    \caption{Comparing confusion matrices for FauxNet, LAA-Net \cite{Nguyen_2024_CVPR}, EfficientNetB4 \cite{9412711} and VideoMAE \cite{tong2022videomae} trained to classify different deepfake generation techniques in proposed Authentica-Vox dataset.}
    \label{confusionMatrixAllvox}
\end{figure*}

\begin{figure*}[h!]
    \centering
    \includegraphics[width=1.0\linewidth]{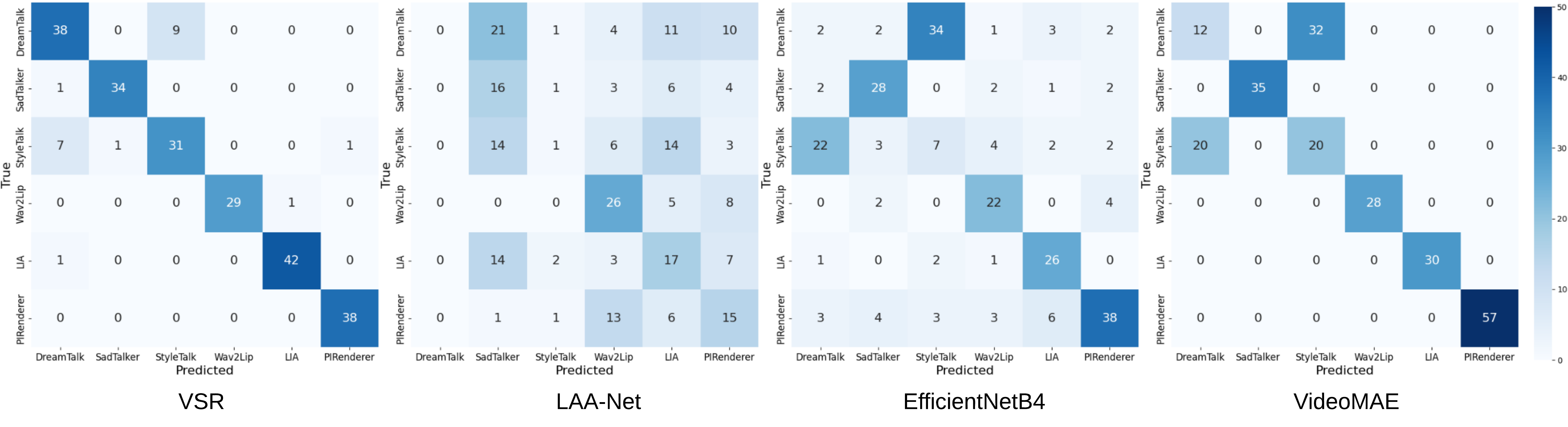}
    \caption{{Comparing confusion matrices for FauxNet, LAA-Net \cite{Nguyen_2024_CVPR}, EfficientNetB4 \cite{9412711} and VideoMAE \cite{tong2022videomae} trained to classify different deepfake generation techniques in proposed Authentica-HDTF dataset.}}
    \label{confusionMatrixAllhdtf}
\end{figure*}

To the best of our knowledge, this is a new perspective in the field. Towards evaluating the accuracy of our model, we extract features from VideoMAE \cite{tong2022videomae}, LAA-Net \cite{Nguyen_2024_CVPR}, and EfficientNetB4 \cite{9412711} and train a classification module in the same multitask framework as our proposed model for fair comparison. In evaluating the accuracy of deepfake attribution, results in Table~\ref{tab:combined_method_comparison} suggest that our model significantly outperforms the rest of the methods. Furthermore, {}{Figures \ref{confusionMatrixAllvox} and \ref{confusionMatrixAllhdtf}} illustrates that the other models are challenged in distinguishing between techniques using the same renderer, \textit{i.e.,}  PiRenderer, StyleTalk and DreamTalk. This showcases the importance of having a distinctly clustered feature space.



\begin{table}[h]
    \centering
    \caption{{{Attribution of deepfakes:} Comparison of FauxNet and {SOTA} on Authentica-Vox and Authentica-HDTF}}
    \begin{tabular}{|c|c|c|}
        \hline
         & \multicolumn{2}{c|}{\textbf{Accuracy (\%)}} \\ 
        \hline
       \textbf{Method} & \textbf{Authentica-Vox} & \textbf{Authentica-HDTF} \\
        \hline
        EfficientNet B4 \cite{9412711} & 56.23 & 52.56 \\
        \hline
        LAA-Net \cite{Nguyen_2024_CVPR} & 48.21 & 32.19 \\
        \hline
        VideoMAE \cite{tong2022videomae} & 76.12 & 77.78 \\
        \hline
        \textbf{FauxNet (ours)} & \textbf{93.35} & \textbf{90.99} \\
        \hline
    \end{tabular}
    \label{tab:combined_method_comparison}
\end{table}

\subsection{Qualitative comparison of feature extractors}
Towards exploring the results obtained on the attribution of deepfakes, we visualize the t-SNE \cite{vandermaaten2008tsne} plots of different feature extractors in {Figures~\ref{fig:tsne_vox} and \ref{fig:tsne_hdtf}}. We clearly observe the separability of feature clusters for different deepfake generation techniques and real videos. Firstly, we verify our hypothesis that the VSR encoder produces considerably distinct features for real and fake videos. In addition, embeddings are clustered based on generation techniques. Interestingly, videos reconstructed from Wav2Lip \cite{prajwal2020lipsync} are clustered {separately} from other techniques and real videos. This is due to Wav2Lip reconstructing only the lip {region}, without {modifying}  {the rest of the face}, \textit{e.g.,} head and pose. Furthermore, PiRenderer-based techniques, LIA, and SadTalker have separable clusters of features, yet they are placed closely as the lip region is influenced by similar head motion and expression traits. 
Although the VSR encoder is distinctly able to separate features of real from fake, {Figure~\ref{fig:tsne_vox} and \ref{fig:tsne_hdtf}} shows that VideoMAE \cite{tong2022videomae}, EfficientNet \cite{9412711} and  LAA-Net \cite{Nguyen_2024_CVPR} are not able to do so. Even though VideoMAE is trained to obtain highly informative features, it does not extract information, instrumental for detecting deepfakes. 
Moreover, interestingly the EfficientNet and LAA-Net are trained with the objective of deepfake detection, yet do not create a distinctly separable feature space for different deepfake generation techniques. This confirms the above-associated quantitative results. 

\subsection{Ablation study}

\begin{table}[h]
    \centering
    \caption{{One versus All Testing on Authentica-HDTF: Comparison of FauxNet and VideoMAE, where only one type of fake generation method is used for training and is evaluated on the test set containing all generation methods.}}
    \begin{tabular}{|c|c|c|c|c|}
        \hline
        \textbf{Train Set} & \multicolumn{2}{c|}{\textbf{FauxNet}} & \multicolumn{2}{c|}{\textbf{VideoMAE}} \\
        \hline
         & \textbf{Acc.\%} & \textbf{AUC} & \textbf{Acc.\%} & \textbf{AUC} \\
        \hline
        Real \& LIA      & 99.67 & 1      & 34.28 & 0.749    \\
        \hline
        Real \& PiRenderer       & 80.20  & 0.9999 & 46.26 & 0.9135   \\
        \hline
        Real \& DreamTalk   & 91.00   & 1      & 79.79 & 0.9693   \\
        \hline
        Real \& SadTalker     & 73.00   & 0.9999 & 48.20  & 0.9147   \\
        \hline
        Real \& StyleTalk       & 85.76 & 1      & 81.44 & 0.9707   \\
        \hline
        Real \& Wav2Lip      & 98.53 & 1      & 26.95 & 0.8068   \\
        \hline
        \textbf{Averaged} & \textbf{88.03} & \textbf{0.9999} & \textbf{52.82} & \textbf{0.8873} \\
        \hline
    \end{tabular}
    \label{tab:combined_one_v_all_hdtf}
\end{table}

\begin{table}[h]
    \centering
    \caption{Ablation on Classification Module on VSR FauxNet {on Authentica-Vox}}
    \begin{tabular}{|c|c|c|}
        \hline
        \textbf{Method} & \textbf{AUC} &\textbf{Accuracy\%}\\
        \hline
        VSR encoder + Linear SVC & 0.9992 &  99.46\\
        \hline
        VSR encoder + GMM Clustering \cite{bishop2006pattern} & 0.9484 & - \\
        \hline
        \textbf{FauxNet (ours)} & \textbf{1.0000} & \textbf{99.86}\\
        \hline
    \end{tabular}
    \label{method_comparison_clf}
\end{table}
\subsubsection{Classification Module}
{Here, we} experiment with classification modules {used to exploit} the separability of real and fake video clusters for VSR encoding features. Table~\ref{method_comparison_clf} summarizes that a combination of Common MLP and Binary Head outperforms the alternative Linear Support Vector Classifier (SVC) and Gaussian Mixture Model Clustering \cite{bishop2006pattern}. This validates our choice for the architecture of the classification module. {Interestingly, the trade-off in using Linear SVC or a GMM model is reasonable enough for real-world applications. Applications requiring a high level of interpretability of the classification module can benefit from the use of Linear SVC \cite{lipton2018mythos}. Similarly, since GMM clustering inherently provides a confidence level \cite{lim2005application} with its prediction, it can be useful in cases where optimizing recall is essential.}

\subsubsection{Classifying features versus text}
Both models are trained and evaluated on the proposed Authentica{-Vox} dataset. Results in Table \ref{method_comparison_text} suggest that FauxNet outperforms the VSR text clustering model. {Apart from the quantitative disadvantage of VSR text clustering, it also requires the ground truth transcript. In case of unavailability, we need to employ speech-to-text, assuming audio has not been manipulated.} This poses a question about the practicality of the model, highlighting the benefits of our proposed FauxNet.

\subsubsection{{Generalizing FauxNet from one seen deepfake generation techniques to other unseen ones}}
{To showcase the generalizing ability of FauxNet, we trained it on several combinations of real + generated data pertained by a single deepfake method. We then test those combinations on the test set, containing data generated by all deepfake methods. We present the results in Table \ref{tab:combined_one_v_all_hdtf}, in which the results of LAA-Net and EfficientNetB4 are omitted due to poor performance. 
The results show the clear superiority of FauxNet, which outperforms VideoMAE on all combinations with an average accuracy of 88.03\% against 52.82\%.}

\begin{table}[h]
    \centering
    \caption{Ablation on Classification Head {on Authentica-Vox}. Comparison between ensembled VSR extracted text-based metric thresholding versus MLP on VSR embeddings (\textbf{FauxNet}).}
    \begin{tabular}{|c|c|c|}
        \hline
        \textbf{Method} & \textbf{AUC} & \textbf{Accuracy\%} \\ 
        \hline
        VSR (text-decoder) + Metric Thresholding & 0.9285 & 95.46\\
        \hline
        \textbf{FauxNet (Ours)} & \textbf{1} &\textbf{99.86} \\
        \hline
    \end{tabular}
    \label{method_comparison_text}
\end{table}

\section{Conclusions and Future Work}
Deepfakes exemplify the intersection of computer vision and ethics, highlighting both the capabilities and challenges posed by rapidly advancing technologies in generative artificial intelligence.

Towards detecting deepfakes, in this work, we first propose a novel dataset, referred to as Authentica. Deviating from previous datasets, Authentica comprises realistic videos stemming from both, video, as well as audio-driven {SOTA} generative techniques.

In addition, we propose FauxNet, a novel framework for generalizable deepfake detection. We present experiments pertaining to in-distribution, as well as zero-shot deepfake detection. 

Finally, we {attribute deepfakes}. We showcase how the pre-trained VSR encoder produces distinctly separable feature clusters for real videos and videos from different generative techniques. Future work will include a larger version of the proposed Authentica dataset, as well as a real-time version of the detection model, which continually adapts detection and classification modules to new deepfake methods through active learning.\\

\vspace{-6mm}
{\small
\bibliographystyle{ieee}
\bibliography{egbib}
}
\vspace{-5mm}
\begin{IEEEbiography}[{\includegraphics[width=1in,height=1.25in,clip,keepaspectratio]{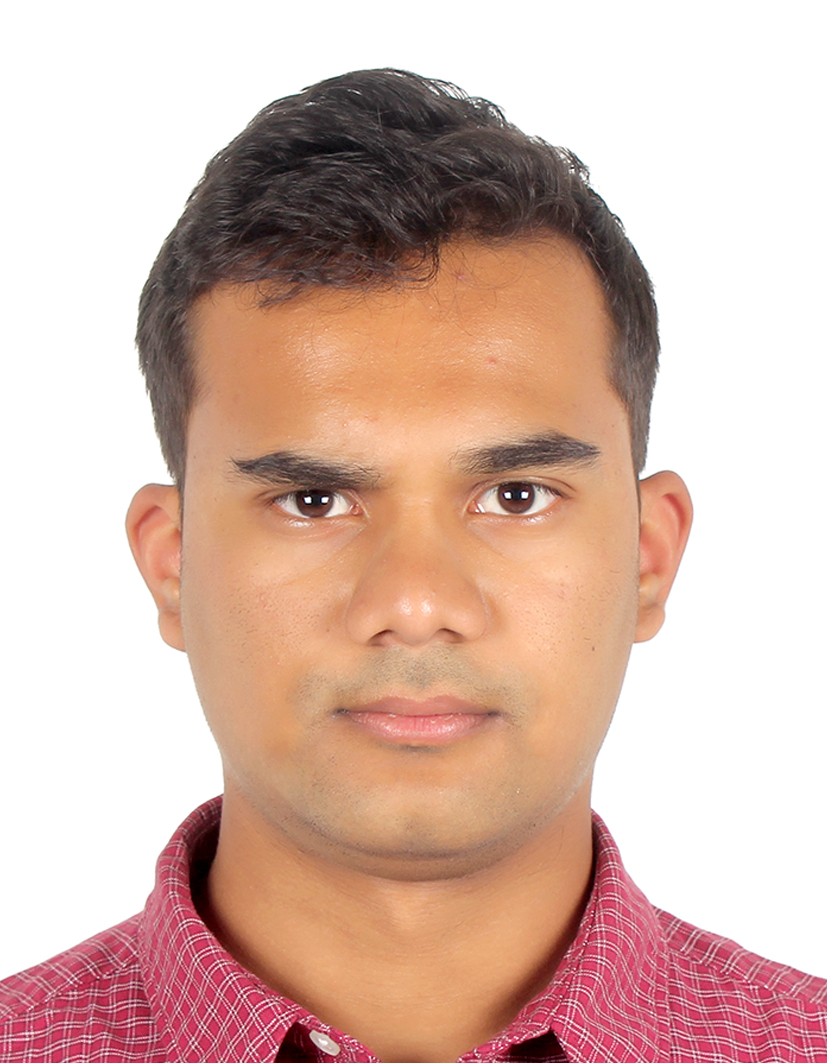}}]{Maheswar Bora} is pursuing a Bachelor of Engineering in Computer Science at BITS Pilani University, Hyderabad campus, India. During his bachelor's, he has pursued his undergraduate thesis as a research intern at under the STARS team at Inria, France. His interests include machine learning and computer vision.
\end{IEEEbiography}

\vspace{-15mm}
\begin{IEEEbiography}[{\includegraphics[width=1in,height=1.25in,clip,keepaspectratio]{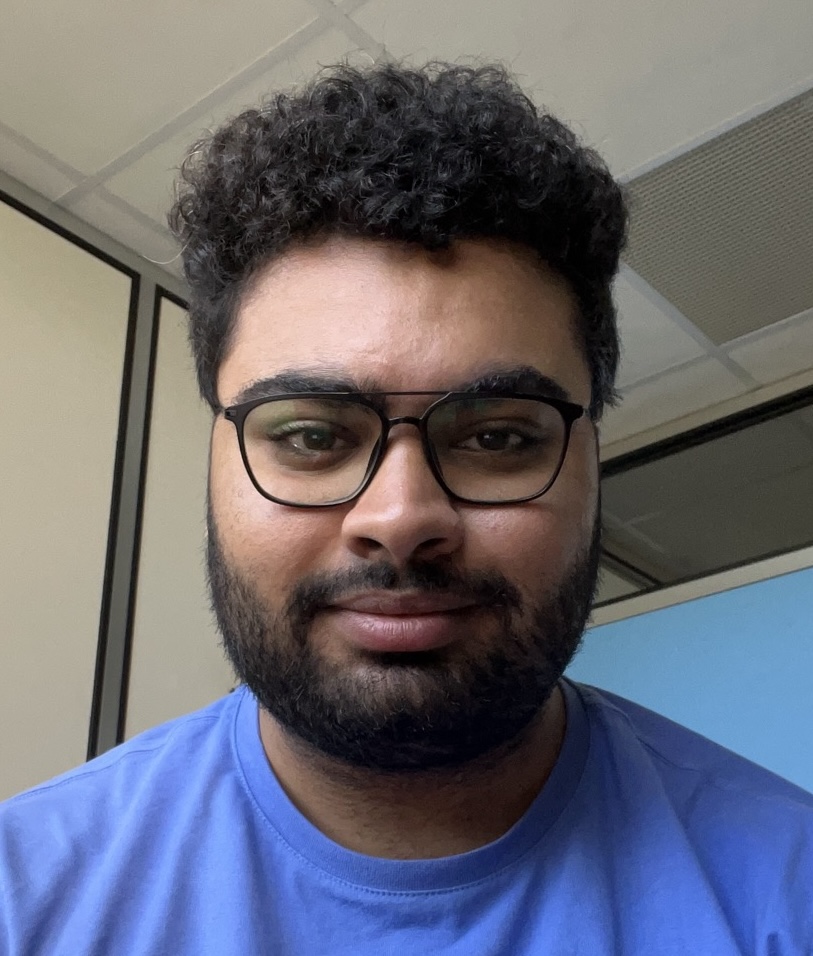}}]{Tashvik Dhamija} is currently pursuing a Master of Science in Artificial Intelligence and Innovation at Carnegie Mellon University, USA. He earned his Bachelor's degree in Electronics and Communications Engineering in 2023 from Delhi Technological University, India. Before pursuing his Masters, he worked as a Computer Vision Engineer and Researcher. His research interests focus on generative vision models, with his current work centred on advancing facial video generation and detection techniques.      
\end{IEEEbiography}

\vspace{-15mm}
\begin{IEEEbiography}[{\includegraphics[width=1in,height=1.25in,clip,keepaspectratio]{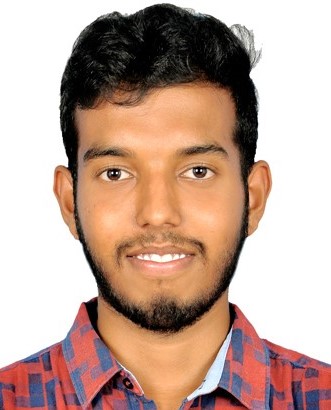}}]{Shukesh Reddy} is currently a PhD scholar in the Department of Computer Science and Information Systems at BITS Pilani, Hyderabad Campus. He earned his Bachelor's degree in Computer Science in 2023. Before pursuing his PhD, he worked as a Software Engineer, leading the development of an IoT platform and IoT development kit. His research interests focus on learning representations of human faces, with his current work centred on advancing face forgery detection techniques.      
\end{IEEEbiography}

\vspace{-15mm}
\begin{IEEEbiography}[{\includegraphics[width=1in,height=1.25in,clip,keepaspectratio]{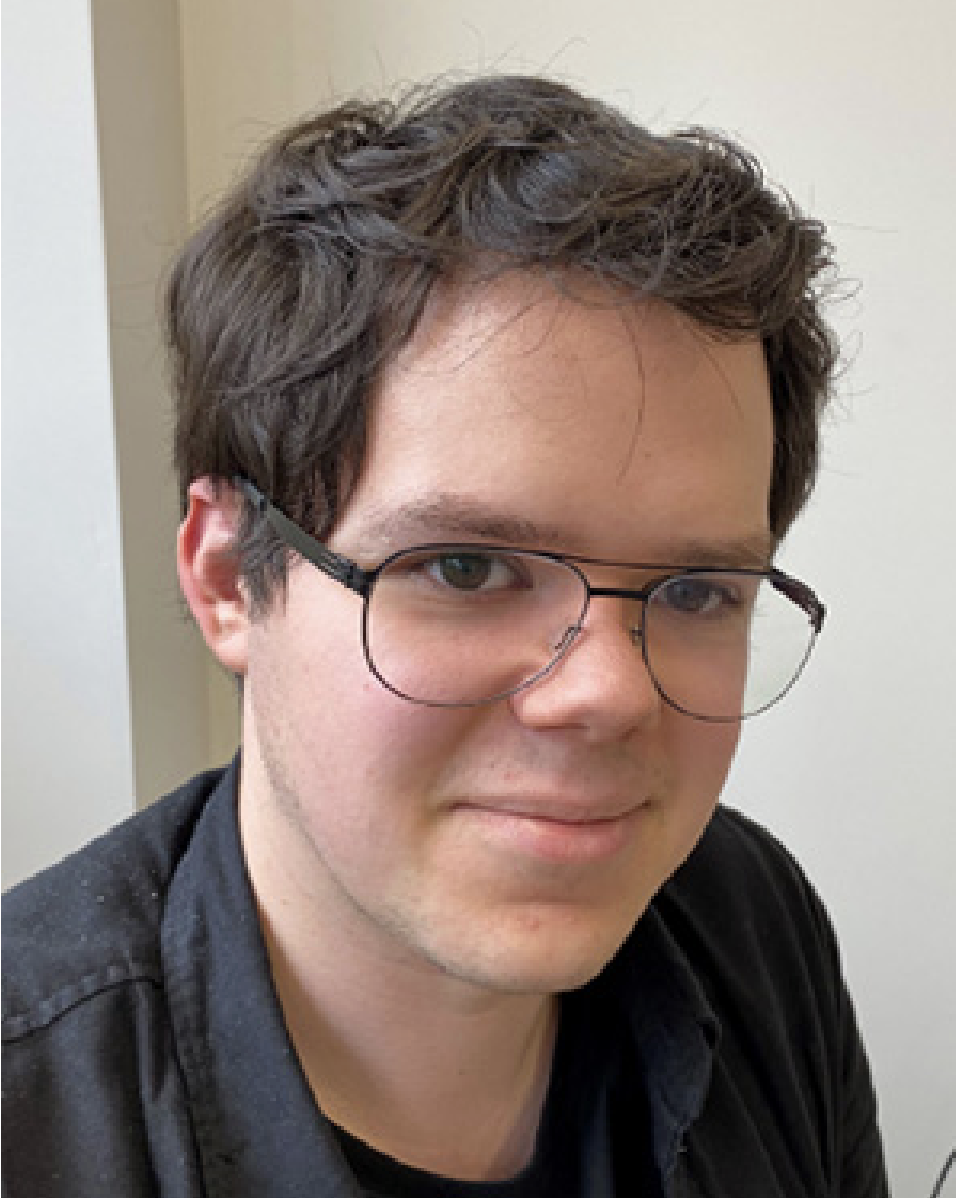}}]{Baptiste Chopin} is currently a Postdoctoral Researcher at Inria, France. Previously, he received his Ph.D. degree at the University of Lille, France and he obtained his engineering degree in computer science from IMT Nord Europe, France. His research concern computer vision and the generation of human motion with application to cognitive sciences.
\end{IEEEbiography}

\vspace{-15mm}
\begin{IEEEbiography}[{\includegraphics[width=1in,height=1.25in,clip,keepaspectratio]{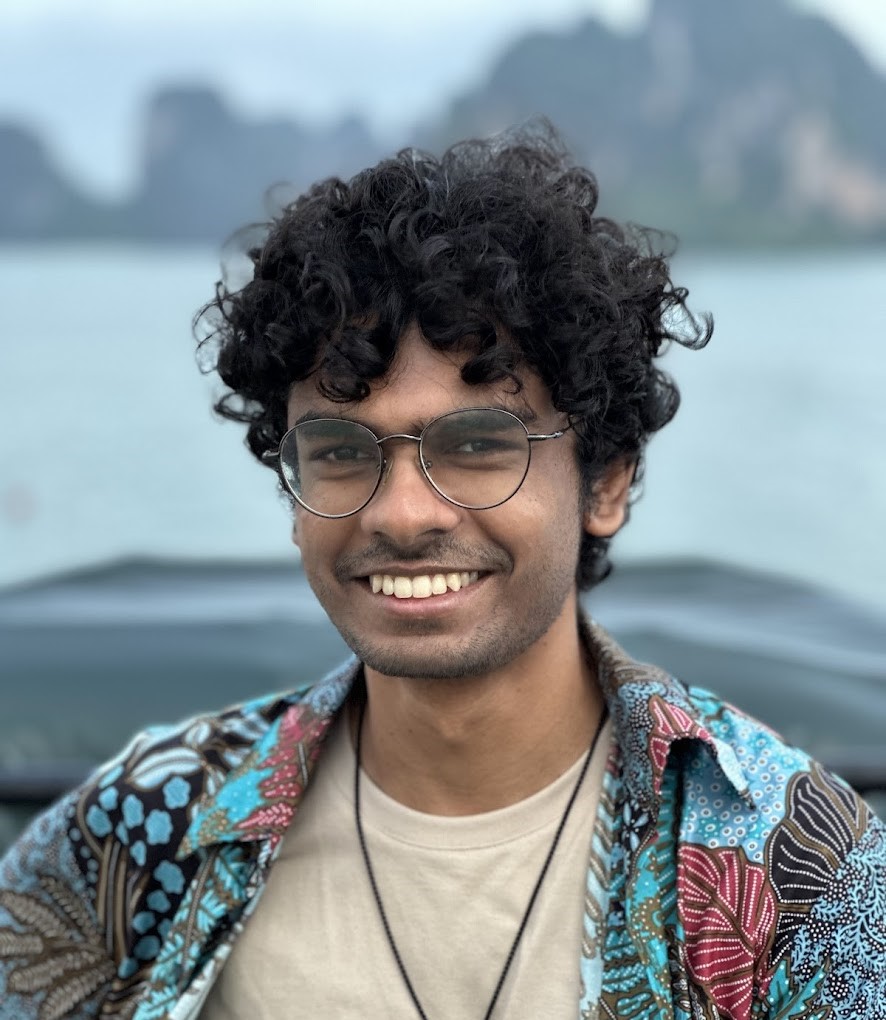}}]{Pranav Balaji} is currently a Computer Vision Data Scientist at Space42, Abu Dhabi. He earned his Bachelor's degree in Computer Science from BITS Pilani, Hyderabad. During his time there, he worked as a Computer Vision researcher and then continued this work in Inria, France. He has worked on Natural Language Processing, Deepfake detection and Generative vision models.

\end{IEEEbiography}

\vspace{-15mm}
\begin{IEEEbiography}[{\includegraphics[width=1in,height=1.25in,clip,keepaspectratio]{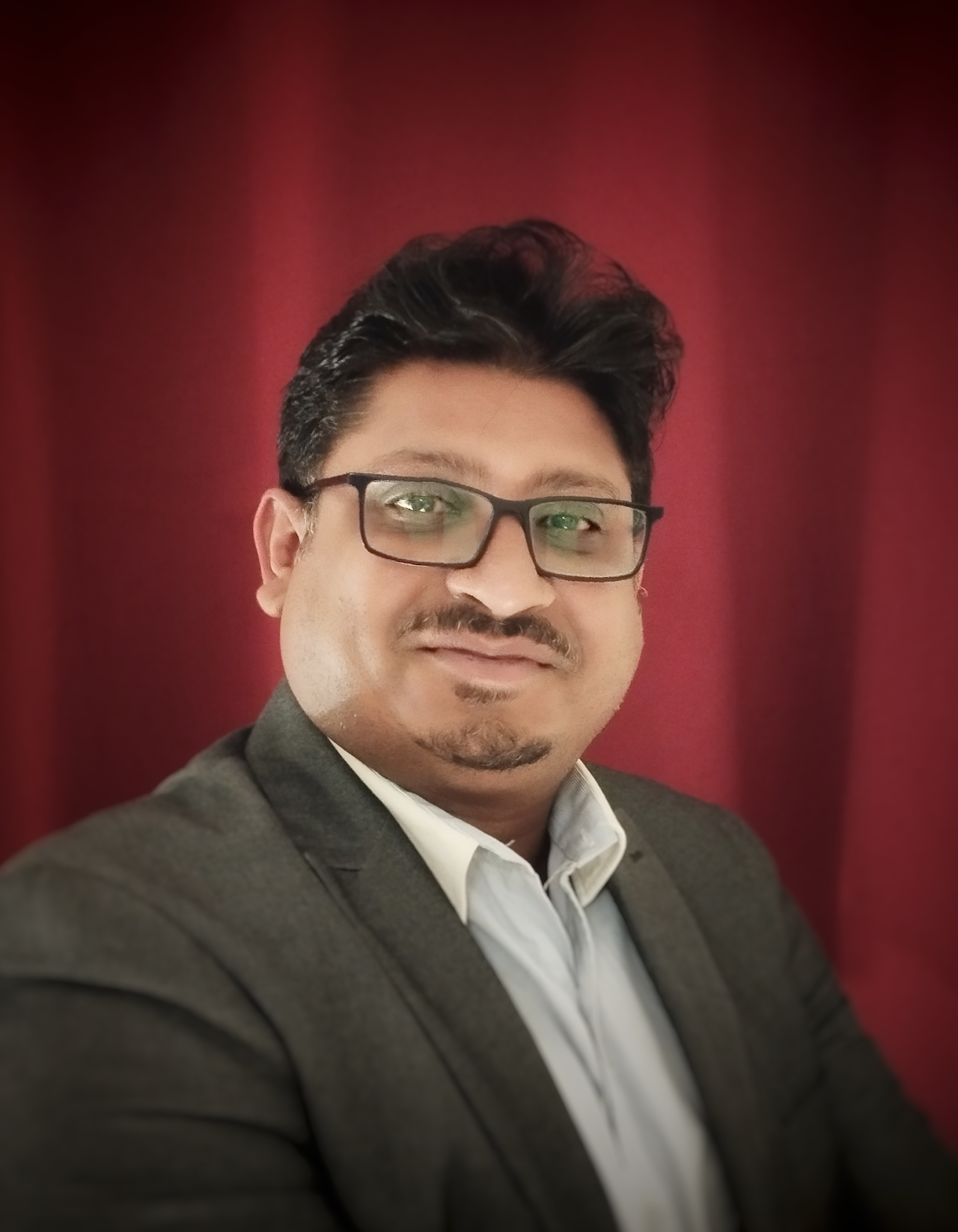}}]{Abhijit Das} is an assistant professor at BITS Pilani Hyderabad. Previously, he worked as a Post-Doc Researcher at Inria Sophia Antipolis– Méditerranée, France. He has completed his PhD from the School of Information and Communication Technology, Griffith University, Australia. He is an accomplished machine learning and computer vision researcher with more than 15 years of research and teaching experience. He is presently pursuing an investigation on learning representations and human analysis employing facial and corporeal-based visual features.
\end{IEEEbiography}

\vspace{-15mm}
\begin{IEEEbiography}[{\includegraphics[width=1in,height=1.25in,clip,keepaspectratio]{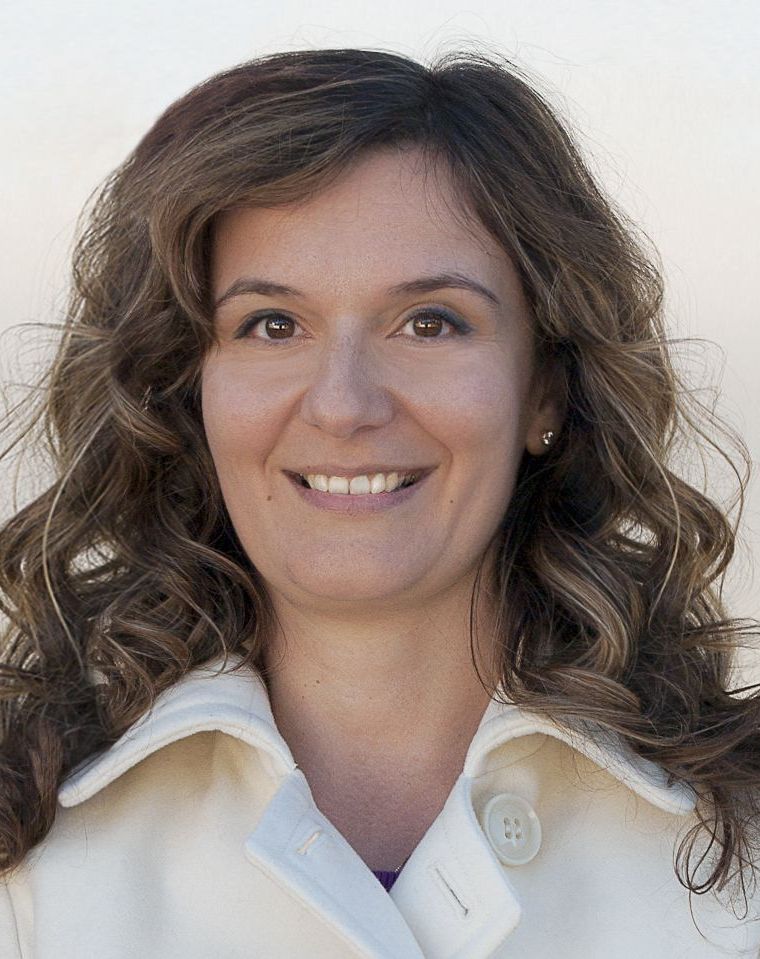}}]{Antitza Dantcheva} Antitza Dantcheva is Directrice de Recherche with the STARS team of Inria Center at Université Côte d'Azur, Sophia Antipolis, France. Previously, she was a Marie Curie fellow at Inria and a Postdoctoral Fellow at the Michigan State University and the West Virginia University, USA. She received her Ph.D. degree from Telecom ParisTech/Eurecom in image processing and biometrics in 2011. Her research is in computer vision and specifically in designing algorithms that seek to learn suitable representations of the human face in interpretation and generation. She is recipient among others of the ANR Jeunes chercheuses / Jeunes chercheurs (JCJC) personal grant, winner of the New Technology Show at ECCV 2022, the Best Poster Award at IEEE FG 2019, winner of the Bias Estimation in Face Analytics (BEFA) Challenge at ECCV 2018 (in the team with Abhijit Das and Francois Bremond) and Best Paper Award (Runner up) at the IEEE International Conference on Identity, Security and Behavior Analysis (ISBA 2017).

\end{IEEEbiography}
\end{document}